\documentclass{article}
\usepackage[T1]{fontenc}
\usepackage{graphicx}
\usepackage[a4paper]{geometry}
\usepackage{fontspec}
\usepackage{todonotes} 
\usepackage{amsmath} 
\usepackage{amssymb}
\usepackage{varioref}
\usepackage{listings}
\usepackage{booktabs}
\usepackage{multirow}
\usepackage{caption}
\usepackage{subcaption}
\usepackage[hidelinks]{hyperref}
\usepackage[capitalize]{cleveref}

\usepackage{makecell}
\usepackage{threeparttable}

\usepackage[dvipsnames]{xcolor}
\usepackage{soul}
\definecolor{keepcolor}{RGB}{115, 197, 231}
\definecolor{replacecolor}{RGB}{255, 215, 128}
\definecolor{insertcolor}{RGB}{243, 185, 185}
\definecolor{deletecolor}{RGB}{243, 185, 185}
\DeclareRobustCommand{\keep}[1]{{\sethlcolor{keepcolor}\hl{#1}}}
\DeclareRobustCommand{\repl}[1]{{\sethlcolor{replacecolor}\hl{#1}}}
\DeclareRobustCommand{\inse}[1]{{\sethlcolor{insertcolor}\hl{#1}}}
\DeclareRobustCommand{\dele}[1]{{\sethlcolor{deletecolor}\hl{#1}}}
\DeclareRobustCommand{\neut}[1]{{{#1}}}

\newcommand{\TP}{\texttt{TP}}
\newcommand{\FP}{\texttt{FP}}
\newcommand{\FN}{\texttt{FN}}

\definecolor{mygreen}{rgb}{0,0.6,0}
\definecolor{mygray}{rgb}{0.5,0.5,0.5}
\definecolor{mymauve}{rgb}{0.58,0,0.82}
\definecolor{lightgrey}{rgb}{0.99,0.99,0.99}

\definecolor{otherlightergrey}{RGB}{247,250,252}
\definecolor{answergrey}{RGB}{207,210,212}
\definecolor{lightergrey}{RGB}{250,250,250}
\definecolor{mediumgrey}{RGB}{185,185,185}
\definecolor{mygreenblue}{RGB}{77,160,187}
\definecolor{mylightblue}{RGB}{74,181,229}
\definecolor{myyellow}{RGB}{206,170,75}
\definecolor{myred}{RGB}{229,105,108}

\definecolor{halfgray}{gray}{0.55}
\definecolor{ipython_frame}{rgb}{0.99,0.99,0.99}
\definecolor{coolgrey}{RGB}{48, 64, 68}

\lstdefinestyle{raw}{
  language=,
  moredelim={[is][\color{mygreenblue}]{@@}{@@}}, % for input
  backgroundcolor=\color{otherlightergrey},   % choose the background color
  basicstyle=\color{coolgrey}\ttfamily,        % size of fonts used for the code
  breaklines=true,                 % automatic line breaking only at whitespace
  captionpos=b,                    % sets the caption-position to bottom
  escapeinside={\%*}{*)},          % if you want to add LaTeX within your code
  rulecolor=\color{lightergrey},
  numbers=none,
  numberstyle=\scriptsize\color{halfgray}\ttfamily,
  numbersep=15pt,
  xleftmargin={0cm},
  frame=single,
  framerule=0pt,
  literate=
  {æ}{{\ae}}1 
  {Æ}{{\AE}}1 
  {ø}{{\o}}1 
  {Ø}{{\O}}1 
  {å}{{\aa}}1 
  {Å}{{\AA}}1 
  {–}{{\textendash}}1 
  {²}{{$^{\tt 2}$}}1 
  {³}{{$^{\tt 3}$}}1
  {<=}{{\(\leq\)}}1,
  showspaces=false,
  showstringspaces=false,
}

\lstdefinestyle{python}{language=python,
  moredelim={[is][\color{mygreenblue}]{@@}{@@}}, % for strengebiter
  backgroundcolor=\color{lightergrey},   % choose the background color
  basicstyle=\color{coolgrey}\ttfamily,        % size of fonts used for the code
  breaklines=true,                 % automatic line breaking only at whitespace
  captionpos=b,                    % sets the caption-position to bottom
  commentstyle=\color{myyellow},    % comment style
  escapeinside={(*@}{@*)},          % if you want to add LaTeX within your code
  keywordstyle=\color{myred},       % keyword style
  stringstyle=\color{mygreenblue},     % string literal style
  rulecolor=\color{lightergrey},
  numbers=left,
  numberstyle=\scriptsize\color{halfgray}\ttfamily,
  numbersep=15pt,
  xleftmargin={0cm},
  frame=single,
  framerule=0pt,
  literate=
  {æ}{{\ae}}1 
  {Æ}{{\AE}}1 
  {ø}{{\o}}1 
  {Ø}{{\O}}1 
  {å}{{\aa}}1 
  {Å}{{\AA}}1 
  {–}{{\textendash}}1 
  {²}{{$^{\tt 2}$}}1 
  {³}{{$^{\tt 3}$}}1
  {0}{{{\ProcessDigit{0}}}}1
  {1}{{{\ProcessDigit{1}}}}1
  {2}{{{\ProcessDigit{2}}}}1
  {3}{{{\ProcessDigit{3}}}}1
  {4}{{{\ProcessDigit{4}}}}1
  {5}{{{\ProcessDigit{5}}}}1
  {6}{{{\ProcessDigit{6}}}}1
  {7}{{{\ProcessDigit{7}}}}1
  {8}{{{\ProcessDigit{8}}}}1
  {9}{{{\ProcessDigit{9}}}}1,
  morestring=[b]',
  morestring=[b]',
  morecomment=[l]//,
  otherkeywords={self, {*}, {=}, {+}, {-}},
  showspaces=false,
  showstringspaces=false,
}

\newcommand\digitstyle{\color{mygreen}}
\makeatletter
\newcommand{\ProcessDigit}[1]
{%
  \ifnum\lst@mode=\lst@Pmode\relax%
   {\digitstyle #1}%
  \else
    #1%
  \fi
}
\makeatother
\lstnewenvironment{python}{\lstset{
  style=python
}}{}
\newcommand{\inpython}[1]{\lstinline[style=python]{#1}}

\newenvironment{monospace}{\parindent=0pt\ttfamily\vspace{0.5em}}{\vspace{0.5em}\\}

\begin{document}
\title{Stringalign: Moving beyond summary statistics with a transparent Unicode-aware tool for evaluating automatic transcription models}
%
%\titlerunning{Stringalign: A tool for evaluating automatic transcription models}
\author{Yngve Mardal Moe\thanks{The authors contributed equally} \thanks{Independent researcher} \and Marie Roald\footnotemark[1] \thanks{The National Library of Norway}}
%\authorrunning{Y. M. Moe and M. Roald}
%\institute{Independent Researcher, Oslo, Norway \and The National Library of Norway, Oslo, Norway \email{marie.roald@nb.no} \\ \textsuperscript{*} Both authors contributed equally to this paper}
%
\maketitle              % typeset the header of the contribution
\begin{abstract}
Comparing text strings is crucial when evaluating and understanding the performance of various text processing tasks such as document recognition and audio transcription. With an increasingly complex landscape of AI-based handwritten text recognition (HTR), optical character recognition (OCR) and automatic speech recognition (ASR) models, there is a need for tools that facilitate evaluation in a flexible and reproducible way. This paper presents Stringalign, a Python library designed to simplify the evaluation process for automatic transcription projects and facilitate transparent evaluation. Stringalign's tools to examine and visualise both the rate of errors and the types of errors a model makes, give insights into possible improvements and help inform model selection for a particular task. Widely used string comparison metrics, such as the character and word error rates (CER and WER), although useful, can be ambiguous due to varying definitions of what constitutes a character and a word. Stringalign addresses this challenge by ensuring all preprocessing (i.e. normalisation and tokenisation) is transparent and easily replicable, and by providing tools to move beyond summary statistics and analyse common model errors. Moreover, Stringalign adheres to FAIR (Findable, Accessible, Interoperable, and Reusable) principles for research software while staying lightweight and easy to adapt into researchers existing workflows. In this paper, we discuss challenges with character and word level string comparisons and show through examples that where existing tools can yield opaque and sometimes confusing results, Stringalign provides an easy-to-use and unambiguous alternative.
%\keywords{Evaluation \and OCR \and HTR \and ATR \and Open-source software}
\end{abstract}
\section{Introduction}\label{sec:introduction}
Machine-learning-based approaches have become increasingly important for document recognition~\cite{11303599}. Therefore, all aspects of document recognition require quality evaluation of model performance. End users need evaluation methods to make an informed choice on which model is right for their use case and researchers need evaluation methods to guide their work in developing new models. Even the data annotation phase of a project may want to evaluate and compare annotations to ensure quality data is produced~\cite{zha2023data}. Tools that enable such evaluation are therefore essential for the entire document recognition stack. 

Similarly, the emergence of deep learning has cemented the importance of data when creating transcription models. Previous work has shown that increasing the size and quality of data is an effective strategy for improving model performance compared to algorithmic optimisations~\cite{10.1145/2347736.2347755,zha2023data}. Still, collecting data can be expensive and time-consuming. Particularly for domains where data is scarce or when the annotation requires domain experts who have little time to devote to annotation, e.g. for low-resource languages~\cite{enstad-etal-2025-comparative,wiechetek-etal-2024-ethical}. For such cases, precise knowledge about what a model struggles with can give insights into what data is lacking so that the data gathering can be planned strategically. 

A critical challenge with evaluating automatic text recognition (ATR) and automatic speech recognition (ASR) models is the ambiguities inherent for comparing strings~\cite{neudecker2021survey}. Normalisation and tokenisation are applied before comparing strings and there are multiple strategies to choose from~\cite{moran2018unicode,santos2019ocr,unicode-annex-15,unicode-annex-29}. These decisions affect metrics such as the character error rate (CER) and the word error rate (WER), yet they are not always reported together with the metrics, making the evaluations hard compare across projects~\cite{neudecker2021survey}. Thus, there is a need for open-source and easy-to-use tools that promotes reproducible evaluation practices.

In later years, the value of research software has been increasingly highlighted. In particular, through Findable Accessible Interoperable and Reusable for Research Software (FAIR4RS) principles which extend open science practices to cover research software \cite{barker2022introducing}. Similarly to the FAIR principles for research data \cite{wilkinson2016fair}, the intent of the FAIR4RS principles is to make sure research software is discoverable, easy to access, compatible with other projects and organised in a way that is easy to reuse. Software aiming to facilitate reproducible and transparent transcription evaluation for research should adhere to these principles.

This work contributes an updated review of the existing transcription evaluation landscape expanding on \cite{neudecker2021survey} and introduces Stringalign which provides:

\begin{itemize}
    \item Well documented and extensively tested code for transcription evaluation that is both in line with FAIR4RS and easy to adapt into existing workflows.
    \item Transparent and Unicode-aware tokenisation.
    \item Tools to explore and visualise model errors.
\end{itemize}

The paper is structured such: \cref{sec:background} discusses challenges of string
comparisons, gives an overview of transcription evaluation tools and motivations for a new tool. \cref{sec:DesignPrinciplesAndImplementation} covers Stringalign's design principles and implementation details. \Cref{sec:examples} contains examples comparing existing tooling and demonstrating Stringalign's features. Finally, \cref{sec:conclusion} contains conclusions and future work.

\section{Background}\label{sec:background}
\subsection{Alignment}
A common approach for comparing strings is through \emph{optimal string alignment}, where two strings, which we denote as the \emph{reference string} and the \emph{predicted string}, are aligned to minimise the number of \emph{edits} required to turn the reference string into the predicted string. Consider the two strings

\begin{monospace}
Reference: string \\
Predicted: align
\end{monospace}
Using the edit operations \texttt{Insert}, \texttt{Delete} and \texttt{Replace}, we can transform the reference into the predicted string: 1. \texttt{Delete(s)}, 2. \texttt{Replace(t, a)}, 3. \texttt{Replace(r, l)}, 4. \texttt{Keep(i)}, 5. \texttt{Delete(n)}, 6. \texttt{Keep(g)}, 7. \texttt{Insert(n)}; or, visualised:

\begin{monospace}
Reference: \dele{s}\repl{tr}\keep{i}\dele{n}\keep{g}\neut{-} \\
Predicted: \neut{-}\repl{al}\keep{i}\neut{-}\keep{g}\inse{n}
\end{monospace}
Fortunately, previous work has defined several efficient algorithms to find such optimal alignments~\cite{navarro_guided_2001,needleman1970general,UKKONEN1985100}.
%A key question then is how do we find the alignment that has the fewest edit operations?
%Luckily, several efficient algorithms solves this problem~\cite{needleman1970general,navarro_guided_2001}.
%Stringalign uses the \emph{Needleman-Wunsch} algorithm~\cite{needleman1970general}, which has \(\mathcal{O}(mn)\), space- and memory-complexity where \(m\) is the length of the reference string and \(n\) is the length of the predicted string. It is, therefore, faster to align many short strings than one long string.

\subsection{Comparison metrics}
\label{sec:comparison_metrics}
\subsubsection{Levenshtein distance}
One metric for comparing two strings is the \emph{Levenshtein distance} \cite{levenshtein1965binary,levenshtein1966binary} (as referenced in \cite{navarro_guided_2001}), also called the \emph{edit distance}\footnote{More precisely, the Levenshtein distance is the edit distance when the edit operations are insertions, deletions and replacements.}. The Levenshtein distance, \(L\), between the reference and predicted strings, \(r\) and \(p\), is
\begin{equation}
    L(r, p) = N_I(r, p) + N_D(r, p) + N_R(r, p),    
\end{equation}
where \(N_I(r, p)\), \(N_D(r, p)\) and \(N_R(r, p)\) is the number of insertions, deletions and replacements in the optimal alignment between \(r\) and \(p\). Thus, \(L(r, p)\) is low for similar strings.
If we consider the example above, we see that the Levenshtein distance between \texttt{string} and \texttt{align} is equal to 5.

\subsubsection{Character Error Rate}
The CER is a common metric for comparing strings, and is given by the number of inserted, deleted and replaced characters divided by the number of characters in the reference string. Specifically, the CER between the reference string, \(r\), and predicted string, \(p\), is given by

\begin{equation}
\text{CER}(r, p) = \frac{N^{(\text{c})}_I(r, p) + N^{(\text{c})}_D(r, p) + N^{(\text{c})}_R(r, p)}{N^{(\text{c})}_D(r, p) + N^{(\text{c})}_R(r, p) + N^{(\text{c})}_K(r, p) }, \label{eq:cer.def}
\end{equation}
where \(N_I^{(\text{c})}(r, p)\), \(N_D^{(\text{c})}(r, p)\), \(N_R^{(\text{c})}(r, p)\) and \(N_K^{(\text{c})}(r, p)\) is the number of inserted, deleted, replaced and kept characters in the optimal character-alignment for \(r\) and \(p\). If we again consider the example above, we see that the CER is \(\frac{5}{6} = 0.83.\)

\subsubsection{Word Error Rate}
Another common metric for comparing strings is the WER, which is computed from an optimal alignment of \emph{words}:

\begin{equation}
\text{WER}(r, p) = \frac{N^{(\text{w})}_I(r, p) + N^{(\text{w})}_D(r, p) + N^{(\text{w})}_R(r, p)}{N^{(\text{w})}_D(r, p) + N^{(\text{w})}_R(r, p) + N^{(\text{w})}_K(r, p) }, \label{eq:wer.def}
\end{equation}
where \(N_I^{(\text{w})}(r, p)\), \(N_D^{(\text{w})}(r, p)\), \(N_R^{(\text{w})}(r, p)\) and \(N_K^{(\text{w})}(r, p)\) are the number of insertions, deletions, replacements and keeps in an optimal word-alignment for \(r\) and \(p\).
Thus, the WER measures the amount of word errors in the predicted string compared to the reference. With the example above, we see that the WER is 1 as both strings contain just one word, which differs.

\subsubsection{Token error rate}

From \cref{eq:cer.def,eq:wer.def}, we see that the WER and CER are similar and the only difference is whether they are computed based on characters or words. A natural generalisation of these concepts is the \emph{token error rate} (TER). To compute the TER, we split the reference and predicted strings into \emph{tokens} (e.g. characters or words) before alignment. Thus, the TER is defined by

\begin{equation}
\text{TER}(r, p) = \frac{N^{(\text{t})}_I(r, p) + N^{(\text{t})}_D(r, p) + N^{(\text{t})}_R(r, p)}{N^{(\text{t})}_D(r, p) + N^{(\text{t})}_R(r, p) + N^{(\text{t})}_K(r, p) },
\end{equation}
where \(N_I^{(\text{t})}(r, p)\), \(N_D^{(\text{t})}(r, p)\), \(N_R^{(\text{t})}(r, p)\) and \(N_k^{(\text{t})}(r, p)\) are the number of inserted, deleted, replaced and kept tokens in an optimal alignment between the tokens of \(r\) and \(p\).
A benefit of the token concept is that characters and words are (as we will discuss in \cref{sec:comparison_ambiguities}) ambiguously defined. Moreover, by considering tokens, we can define metrics that are valid for both words and characters.

\subsubsection{Token-specific metrics}
For some applications it can be useful to look at performance metrics for specific tokens~\cite{enstad-etal-2025-comparative,rice1996isri}.  % maarand2022comprehensive
Based on the optimal alignment, we can, e.g. compute the true positive count (\TP), false positive count (\FP) and false negative count (\FN) for each token in the predicted and reference strings. In particular, we count the number of times it is kept (\TP), the number of times it is inserted or replaces another token (\FP) and the number of times it is deleted or replaced with another token (\FN).
From these numbers, we can compute a variety of token-specific statistics, which are defined in \cref{tab:token_specific_metrics}.

\begin{table}[t]
    \centering
    \caption{Token-specific metrics}
    \begin{tabular}{@{}r@{\hspace{1em}}r@{ = }l@{\hspace{1em}}l@{}}
    \toprule
        Metric & \multicolumn{2}{l}{Definition} & Synonyms \\
    \midrule
        True positive rate & \(\text{TPR}(t)\) & \(\frac{\TP(t)}{\TP(t) + \FN(t)}\) & Sensitivity, Recall \\
        Positive Predictive Value & \(\text{PPV}(t)\) & \(\frac{\TP(t)}{\TP(t) + \FP(t)}\) & Precision \\
        False Discovery Rate & \(\text{FDR}(t)\) & \(\frac{\FP(t)}{\TP(t) + \FP(t)}\) & \\
        \(f_1\)-score & \(f_1(t)\) & \(2\frac{\text{TPR}(t)\text{PPV}(t)}{\text{TPR}(t) + \text{PPV}(t)}\) & Dice score\\
    \bottomrule
    \end{tabular}
    \label{tab:token_specific_metrics}
\end{table}

\subsection{Alignment uniqueness}\label{sec:uniqueness}

The optimal alignment is not \emph{unique} as two strings may have more than one alignment with the same number of edit operations.
Consider the earlier example:

\begin{monospace}
Reference: \dele{s}\repl{tr}\keep{i}\dele{n}\keep{g}\neut{-} \\
Predicted: \neut{-}\repl{al}\keep{i}\neut{-}\keep{g}\inse{n}
\end{monospace}
Here, the Levenshtein distance is five, but multiple alignments have five edits. In particular, the first deletion has three possible locations (delete \texttt{s}, \texttt{t}, or \texttt{r}), the transposition of \texttt{n} and \texttt{g} can be resolved in three ways (aligning the \texttt{n}s, \texttt{g}s or neither), and the first deletion can be omitted, aligning \texttt{stri} with \texttt{alig}, keeping the \texttt{n} and deleting the \texttt{g}. In total, we have ten different optimal alignments where the TP, FP and FN for each token can vary, while the TER and Levenshtein distance are constant.
Moreover, the number of possible optimal alignments can grow quickly with string length. 
Still, character confusions and token-specific statistics can provide useful information on what tokens a transcription model struggles with, and the non-uniqueness challenge can be alleviated by averaging the \TP, \FP \ and \FN \ for several different optimal alignments.

Another way to mitigate this non-uniqueness challenge is merging sub\-sequent edits. Then, the above example would get the multi-token edit \texttt{str -> al}, reducing the number of possible alignments to four. This strategy makes counting common edit operations more stable, but makes \TP, \FP \ or \FN \ ill-defined.
%However, in practice, this problem is less severe, as similar strings seldom have many different optimal alignments.

\subsection{Ambiguities in string normalisation and segmentation}
\label{sec:comparison_ambiguities}
So far, we have glossed over some questions critical for comparing strings in practice: What constitutes a ``character''? And when are two characters ``different''? This section discusses how characters are represented in a computer and how ambiguities in the representation can lead to ambiguities in evaluation metrics.

\subsubsection{Unicode}
Computers represent characters through enumeration.
Typically, \texttt{a} is represented by the number \(97\), or \texttt{0x61} in hexadecimal, \texttt{b} is 98, or \texttt{0x62}, etc. 
The mapping from numbers (or \emph{code points}) to characters is a \emph{string} \emph{encoding}, and the most common string encoding is Unicode~\cite{santos2019ocr,unicode-standard}.
Unicode code points are often represented as hexadecimal numbers prepended with \texttt{U+}, so \texttt{a} is \texttt{U+0061}.

%The Unicode standard defines this mapping between code points and \emph{abstract characters} (see \cref{sec:grapheme.clusters}).
%However, the standard does not specify a single correct way of storing these code points in memory, which is the purpose of the 8-, 16- and 32-bit Unicode Translation Formats (UTF-8, UTF-16 and UTF-32).
%This text assumes that strings are stored as sequences of Unicode code points with the exact memory layout for these sequences being irrelevant.

\subsubsection{Normalisation}
\label{sec:normalisation}
Some characters have more than one Unicode representation~\cite{unicode-annex-15}. Such ambiguities can make it unclear how to count or compare strings. Therefore, Unicode defines two equivalence classes for characters: \emph{canonical equivalence} and \emph{compatibility equivalence} \cite{unicode-annex-15,unicode-standard}.
Canonically equivalent characters are code points, or code point sequences, that represent the same abstract character and should always look the same.
Meanwhile, compatibility equivalent characters do not need to look the same and can be both visually and semantically different.

%\begin{table}
%    \centering
%    \caption{Equivalent characters with different Unicode representations}
%    \begin{tabular}{l@{\hspace{2ex}}l@{\hspace{2ex}}l}
%    \toprule
%        Character   & Code point sequence & Code point names \\
%    \midrule
%        Å                   & \texttt{U+00C5} (Å)                                                 & Uppercase letter Å \\
%        \multirow{ 2}{*}{Å} & \multirow{ 2}{*}{\texttt{U+0041} (A) \texttt{U+030A} (\(~^\circ\))} & Uppercase letter A \\
%                            &                                & Combining Ring Above  \\
%        Å                   & \texttt{U+212B} (Å)                              & Ångstrøm sign   \\
%    \end{tabular}
%    \label{tab:multi.codepoint.char}
%\end{table}

Unicode also defines composed and decomposed forms for each equivalence class. The decomposed form expands characters into as many constituent code points as possible and sorts them, while the composed form then iteratively merges compatible code points to get a compact representation. In total, we get the four Unicode normalisation forms: canonical decomposed (NFD), canonical composed (NFC), compatibility decomposed (NFKD) and compatibility composed (NFKC). These normalisation forms are illustrated in \cref{tab:normalisation}. Since NFC-normalised text is as compact as possible while not modifying the visual appearance, it is a normalisation likely to fit most research applications.

%We need to be careful when deciding if compatibility equivalence is suitable, as the difference between compatibility equivalent forms can in some contexts be more than stylistic. For example, in mathematical notation, different styles can be used to represent different information.

\begin{table}[t]
    \centering
    \caption{Various characters and their different normalisation forms}
    \begin{tabular}{c@{\hspace{2ex}}c@{\hspace{2ex}}c@{\hspace{2ex}}c@{\hspace{2ex}}c}
    \toprule
        Character & NFD & NFC & NFKD & NFKC \\
    \midrule
        Å  & A\(~^\circ\)  & Å & A\(~^\circ\)  & Å \\
        \texttt{U+00C5} & \texttt{U+0041 U+030A} & \texttt{U+00C5} & \texttt{U+0041 U+030A} & \texttt{U+00C5} \\[0.4em]
        %ﬁ  & ﬁ  & ﬁ & fi  & fi \\
        %\texttt{U+FB01} & \texttt{U+FB01} & \texttt{U+FB01} & \texttt{U+0073 U+0074} & \texttt{U+0073 U+0074} \\[0.4em]
        \(\mathfrak{T}\) & \(\mathfrak{T}\) & \(\mathfrak{T}\) & T & T \\
        \texttt{U+1D517} & \texttt{U+1D517} & \texttt{U+1D517} & \texttt{U+0054} & \texttt{U+0054} \\[0.4em]
         ſ̣  & ſ̣  & ſ̣  & ṣ & ṣ \\
        \texttt{U+017F U+0323} & \texttt{U+017F U+0323} & \texttt{U+017F U+0323} & \texttt{U+0073 U+0323} & \texttt{U+1E63} \\
    \bottomrule
    \end{tabular}
    \label{tab:normalisation}
\end{table}

\subsubsection{Confusable characters}
Unicode also contains \emph{confusable characters} or \emph{confusables} (also called homoglyphs or homographs)~\cite{moran2018unicode,unicode-annex-39}. Such characters look the same or similar but are different and cannot be normalised to the same code point. %How similar two confusables appear can depend on the font used (e.g. I/l,\textsf{I}/\textsf{l} and \texttt{I}/\texttt{l}). 
Addressing this, Unicode provides two lists: \texttt{confusables.txt}, containing visual confusables useful for detecting security problems and \texttt{intentional.txt}, containing characters often identical in typefaces with a harmonised design~\cite{unicode-annex-39}. During evaluation, you should consider how you expect models to handle confusables as this can depend on the type of model and application. 

%In particular, the Unicode consortium defines the \texttt{confusables.txt}-list, which contains visual confusables that can be useful for detecting security problems and the \texttt{intentional.txt}-list, which contains characters that that is probably identical in typefaces with a harmonised design~\cite{unicode-annex-39}. 

%In particular, the Unicode consortium defines two lists: \texttt{confusables.txt} and \texttt{intentional.txt}. \texttt{confusables.txt} contains visual confusables that can be useful for detecting security problems, while \texttt{intentional.txt} contains characters often identical in typefaces with a harmonised design~\cite{unicode-annex-39}. 

%During evaluation, you should consider how you expect models to handle confusable characters as this can depend on the type of model and application. 
%A simple, purely visual OCR model, classifying character by character, would maybe not be able to distinguish between identical confusable characters, so it might be appropriate to resolve them before evaluating.
%However, if an OCR model also models language, you might expect the model to be able to choose correctly among confusable characters.

\subsubsection{Character segmentation} \label{sec:grapheme.clusters}
A simple strategy for character segmentation is segmenting on code points, 
but, as we saw in \cref{sec:normalisation}, a character can consist of multiple code points. 
Some such cases can be normalised to one code point, but this is not always possible (e.g. ą́ )~\cite{moran2018unicode,unicode-annex-29}. Thus, segmenting on code points can give unexpected effects when counting tokens, which can affect the CER (or any metric described in \cref{sec:comparison_metrics}). As such, Unicode defines a method for detecting so-called \emph{grapheme cluster boundaries}\footnote{Note that establishing a completely unambiguous definition of user perceived characters is not always possible~\cite{unicode-annex-29}.}~\cite{unicode-annex-29}. Character-level metrics should therefore segment on these boundaries to ensure a standardised result. 

\subsubsection{Word segmentation}
As it is not clear how to treat aspects like different whitespace and punctuation types, word segmentation is also ambiguous. One approach is to split on \texttt{' '}, however, this does not account for other whitespace types (e.g. tabluation). 
A more robust approach is to split at any Unicode whitespace or any Unicode space, segment or paragraph separator.
Alternatively, Unicode defines word boundaries for extracting words without punctuation. 
These strategies work well on many languages using Latin, Arabic or Devanagari scripts, but they do not work sufficiently e.g. for many Asian languages using other scripts~\cite{unicode-annex-29}.

\subsection{Related work}\label{sec:related_work}
A variety of tools for string comparison exists, targeting different needs. Still, each tool has limitations hindering transparent and reproducible research. Here, we examine six such tools and what limitations in the current landscape motivates a new tool. Specifically, we compare whether tools support token-specific metrics, are libraries or applications, have a transparent documentation (explaining how a tool works rather than just how to use it), have Unicode-aware character and word-segmentation, support custom tokenisation and provide alignment visualisation. As a proxy for how easy the tools can be integrated in other workflows, we also compare their disk footprint. \Cref{tab:ToolOverview} provides a brief overview.

The IMPACT centre's ocrevalUAtion tool\footnote{\url{https://github.com/impactcentre/ocrevalUAtion}} and the Information Science Research Insititue (ISRI) analytic tools for optical character recognition (OCR) evaluation \cite{rice1996isri} are applications that create text reports for CER, WER and some character-specific metrics. Neither of these tools address the optimal alignment's non-uniqueness, and both segment characters as code points. While the ISRI analytic tools have been extended with Unicode word segmentation~\cite{santos2019ocr}, ocrevalUAtion use an undocumented regular expression. Moreover, as both tools produce text reports, they can be difficult to integrate in a unified evaluation  pipeline, and out of the two, only the ISRI analytic tools support aggregating statistics across samples.

Calamari~\cite{wick_calamari_2020} is a framework for the full OCR pipeline and can report CER, WER and common edits. However, for the latter, it uses a heuristic based on recursively finding the longest matching substrings, which can result in a suboptimal alignment (and is also not unique). Finally, as it is built for Calamari models, it does not support seamless evaluation of non-Calamari models.

Similarly, Dinglehopper\footnote{\url{https://github.com/qurator-spk/dinglehopper}} is an application for evaluating OCR outputs that integrates with the OCR-D platform~\cite{neudecker2019ocr} and reports CER, WER and a visualisation of the full documents with highlighted edits. Dinglehopper uses grapheme clusters and Unicode word segmentation for characters and words, respectively\footnote{with the \inpython{uniseg} Python Library}. However as it is built to integrate with a comprehensive platform and lacks transparent documentation, it can be difficult to integrate into an arbitrary workflow.

Unlike the other tools covered in this section, Meeteval and Jiwer are geared towards ASR-evaluations. Jiwer is lower-level and provides utilities for aligning strings, computing performance metrics like CER and WER, and visualising alignments in the command line. While Jiwer supports arbitrary tokenisation, it does only minimal preprocessing by default, with code points as characters and segmenting words based on \texttt{' '}. Meeteval, on the other hand, is higher-level, with visualisations and WER-variants aimed at multi-speaker transcriptions. Meeteval's targeted focus means that it only supports word-tokenisation (based on whitespace-like characters) and it is challenging to customise to other needs.

Earlier work compared several of the tools described above and showed that even standardised metrics like CER and WER cannot be directly compared across implementations \cite{neudecker2021survey}. To improve this state, they suggested researchers follow clear OCR evaluation standards and consider provenance, i.e. by keeping track of and reporting evaluation parameters. Analysing the error types, not just the rate has also been shown to give valuable insights~\cite{8791206}. 

\begin{table}[t]
\centering
\begin{threeparttable}
\caption{Comparison of transcription model evaluation tools}
\label{tab:ToolOverview}
\begin{tabular}{l@{\hspace{2ex}}lllllll}
\toprule
                           & \makecell{Calamari\\~}     & \makecell{Dingle-\\hopper} & \makecell{ISRI\\~}   & \makecell{Jiwer~\\~} & \makecell{Meeteval\\~} & \makecell{ocreval-\\UAtion} & \makecell{String-\\align} \\
\midrule
Token-specific metrics\tnote{+}     & Yes\tnote{*}          & No            & Yes\tnote{*}       & No      & No       & Yes\tnote{*}           & Yes           \\
Library or executable      & Exe. & Exe. & Exe. & Both & Both     & Exe. & Lib. \\
Transparent docs.  & No            & No            & Yes         & No       & No        & No             & Yes           \\
Installed size             & 2.3\thinspace G      & 0.41\thinspace G      & 0.73\thinspace M   & 2\thinspace M  & 0.24\thinspace G  & 13\thinspace M        & 83\thinspace M      \\
Unicode-aware              & No           & Yes          & Words~     & No      & No       & No            & Yes         \\
Custom tokenisation        & No           & No           & No        & Yes     & No       & No            & Yes         \\
Alignment vis. & No            & Yes          & No        & Yes     & Yes      & Yes           & Yes\\        
\bottomrule
\end{tabular}
\begin{tablenotes}
    \item[+] The different tools support slightly different token-specific metrics
    \item[*] Does not address the non-uniqueness challenge of optimal alignments
\end{tablenotes}
\end{threeparttable}
\end{table}

\section{Design principles and implementation details}\label{sec:DesignPrinciplesAndImplementation}
\subsection{Design principles}
As discussed, there is a need for easy-to-use tooling for transparent, reproducible evaluation in automatic transcription research. This section outlines how Stringalign is designed to fill this need. First, we describe how Stringaligns API facilitates transparent comparisons. Then, how its modular and low-dependency architecture enables Stringalign to be used across different setups. Finally, as Stringalign is designed for research, we describe how it adheres to FAIR4RS.

\subsubsection{Facilitating reproducible and transparent comparisons}
%\todo[inline]{In this work, we build on this perspective by providing a tool designed to promote standardised and unambiguous evaluation and returns the evaluation parameters like text transforms together with metric scores to make it easy for researchers to include them in reports.}
There are, as mentioned in \cref{sec:background}, many subtle choices that can affect metrics widely used to evaluate automatic transcription models. Motivated by this, Stringalign is, in contrast to other transcription evaluation tools, explicit in its tokenisation and normalisation and base them on standardised Unicode recommendations. Researchers start by defining a tokeniser before comparing strings, or they use utility functions that return the tokeniser and performance metrics. Thus, Stringalign aids in reporting the exact steps needed to reproduce results.

Furthermore, by supporting token-specific metrics, as described in \cref{sec:uniqueness}, Stringalign encourages researchers to move beyond summary statistics and get a deeper insight into model performance. Moreover, to tackle the non-uniqueness challenge of optimal string alignment, Stringalign has the option to randomise alignments, or even iterate over all possible optimal alignments, which can be used to estimate averages and standard deviations for the token-specific metrics. 

\subsubsection{Lightweight architecture}
Research code can have, and often has, a very large number of dependencies. These dependencies can, sometimes, be both necessary and beneficial, but each dependency increases the likelihood of conflicts\footnote{To illustrate this, consider the Calamari OCR framework, which requires TensorFlow installed with a version between \texttt{2.4.0} and \texttt{2.15.x} (inclusive). This limits the applications its evaluation pipeline can be integrated with easily.}.

To aid comparable evaluation across different projects Stringalign is designed to be lightweight and easily fit into various workflows. Effort has been placed into keeping Stringalign’s dependencies minimal, without sacrificing usability. Also, Stringalign is scoped to only cover transcription model evaluation. By limiting Stringalign’s scope, we make the code more forward-compatible, increasing its long-term maintainability. This focus also enables researchers to include Stringalign in almost any Python project without altering existing environments.

\subsubsection{FAIR research software}

The past decade, the importance of good practices for research software has been increasingly acknowledged cumulating in, among others, the FAIR4RS~\cite{barker2022introducing} principles which state that research software should be:
\begin{description}
    \item[Findable:] Software, and its associated metadata, is easy for both humans and machines to find.
    \item[Accessible:] Software, and its metadata, is retrievable via standardised protocols.
    \item[Interoperable:] Software interoperates with other software by exchanging data and/or metadata, and/or through interaction via application programming interfaces (APIs), described through standards.
    \item[Reusable:] Software is both usable (can be executed) and reusable (can be understood, modified, built upon or incorporated into other software).
\end{description}
As Stringalign is intended for research, following these principles are essential.

\paragraph{Findable}
To ensure that Stringalign is findable, it is hosted on GitHub\footnote{\url{https://github.com/yngvem/stringalign/}} with releases posted to PyPI\footnote{\url{https://pypi.org/project/stringalign/}} and Zenodo\footnote{\url{https://zenodo.org/records/18808052}}. Moreover, Stringalign follows the latest standards in the Python ecosystem with respect to software metadata\footnote{E.g. PEP 517 and 518 for build system metadata and PEP 621, 639 and 735 for main software metadata, such as license information, authors, etc.}. By publishing releases on Zenodo and PyPI, we also ensure that the software is archived with a digital object identifier (DOI). Thus, Stringalign and its associated metadata is easy for both humans and machines to find.

\paragraph{Accessible}
By following the packaging standards in the Python ecosystem, String\-align is accessible as both built artifacts (wheels) and source distributions (sdists).
Furthermore, the built binaries are available for many platforms and CPU architectures, increasing accessibility.
These distributions are available via HTTP, ensuring that Stringalign is retrievable via standardised protocols.

\paragraph{Interoperable}
Stringalign supports all Python versions \(\geq\)\texttt{3.11}. Furthermore, Stringalign has only one runtime dependency: NumPy and supports all NumPy versions \(\geq\)\texttt{1.24.0}. Thus, Stringalign supports all Python- and NumPy-versions that the Scientific Python ecosystem suggests\footnote{See SPEC-0: \url{https://scientific-python.org/specs/spec-0000/}}. Also, Stringalign's wheels are built against the stable Python Application Binary Interface (ABI), which makes them forward compatible with respect to future Python and NumPy\footnote{NumPy guarantees a backwards compatible ABI} releases.

Moreover, Stringalign puts minimal assumptions on input format. This choice contrasts with some ATR evaluation tools. By building on pure Python strings instead of specific file and file system structures, we increase interoperability. Still, to support domain-specific formats, Stringalign's documentation shows how to extract text-lines from PAGE and ALTO XML. Consequently, researchers from any domain needing string comparison can easily use Stringalign.

In short, Stringalign is built to interoperate with Python and other libraries through a clearly defined API based on Python and NumPy primitives.

\paragraph{Reusable}
To make it easy to use, reuse and modify, Stringalign has a clearly designed API and a test suite, covering over \(95 \%\) of the code. The intuitive API makes Stringalign user-friendly and the test coverage grants users confidence to adapt Stringalign to their needs without inadvertently breaking functionality.

Stringalign's in-depth documentation\footnote{\label{footnote:docs}Available on \url{https://www.stringalign.com/}} further aids researchers in both using and understanding the tools provided. Inspired by \emph{Diátaxis}\footnote{\url{https://diataxis.fr/}}, it has four main parts: Concepts, examples, a tutorial and API documentation. Concepts cover important topics such as alignment, normalisation, and grapheme clusters; examples contain code snippets that are easy to copy-paste and reuse; the tutorial explains a real-world analysis using Stringalign; and the API documentation describes all public-facing functionality. By helping researchers understand every part of the evaluation, the documentation promotes transparent use of the code.

Thus, Stringalign's transparent documentation, extensive test suite, clear design and permissive open-source MIT license ensures that it can easily be understood, modified, built upon or incorporated into other software.

\subsection{Implementation details}
In this section, we describe the six public-facing modules in Stringalign.

\begin{figure}[t]
    \centering
    \begin{subfigure}[t]{\textwidth}
        \includegraphics[width=\linewidth]{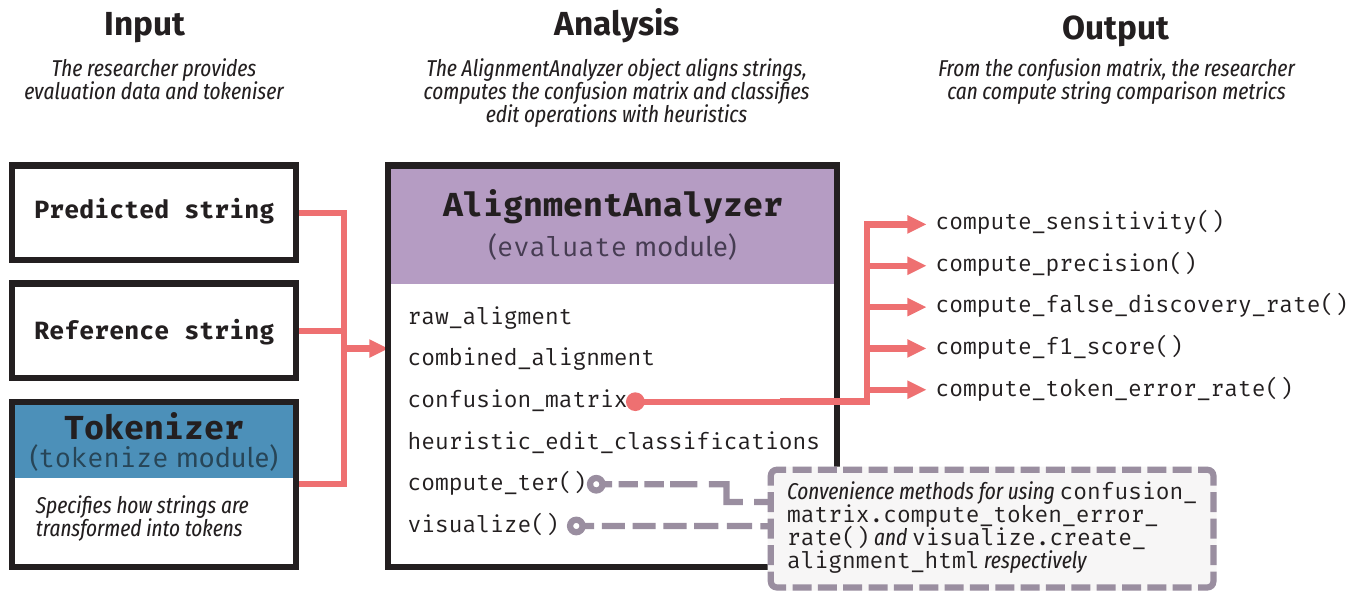}
    \end{subfigure}
    \begin{subfigure}[b]{\textwidth}
        \includegraphics[width=\linewidth]{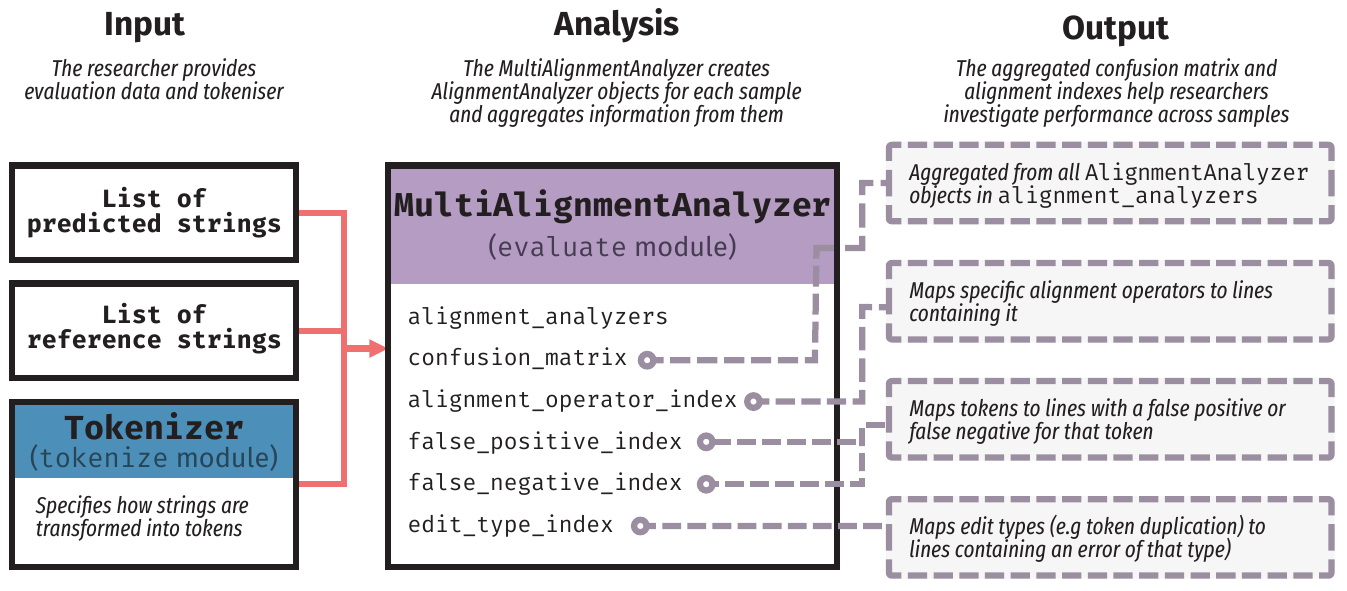}
    \end{subfigure}
    \caption{Inputs and most commonly used attributes, methods and properties for the \inpython{AlignmentAnalyzer} and \inpython{MultiAlignmentAnalyzer} classes}
    \label{fig:CodeFlow}
\end{figure}

\inpython{stringalign.evaluate} is the main module and has two main classes: the \inpython{AlignmentAnalyzer} and the \inpython{MultiAlignmentAnalyzer}. The former has functionality for analysing a single reference/prediction pair such as computing TER, token-specific metrics and heuristic-based edit classification, listing edit freq\-uencies and visualising alignments. The latter aggregates \inpython{AlignmentAnalyzer}s to compute micro-averaged dataset-level metrics. The \inpython{MultiAlignmentAnalyzer} also adds indexes to look up and examine reference/prediction string pairs based on edit operations. See \cref{fig:CodeFlow} for an overview of these classes' main functionality. 

\inpython{stringalign.tokenize} contains tokeniser classes for character and word segmentation. The \inpython{GraphemeClusterTokenizer}, \inpython{UnicodeWordTokenizer} and \inpython{SplitAtWordBoundaryTokenizer} use Stringalign’s bindings to the Rust crate \texttt{unicode\_segmentation}\footnote{\url{https://docs.rs/unicode-segmentation/latest/unicode_segmentation/}}. The \inpython{SplitAtWhitespaceTokenizer} can also tokenise words. All tokenisers support normalising text before and after tokenisation. NFC normalisation is used by default, but other normalisers can be supplied.

\inpython{stringalign.normalize} provides a \inpython{StringNormalizer} class, which can apply Unicode normalisation, case-fold, remove or standardise whitespace, remove non-word characters and resolve confusables using either official Unicode lists or custom mappings. The normaliser also ensures transforms are applied in the correct order. For example, for case-insensitive NFKD normalised comp\-arison, the correct order is case-fold\(\to\)NFKD-normalise\(\to\)case-fold\(\to\)NFKD-normalise~\cite{unicode-standard}. 

\inpython{stringalign.statistics} defines \inpython{StringConfusionMatrix}. This class computes token-specific statistics based on counts of edit operators, \TP, \FP \ and \FN. Researchers will typically not need to instantiate this class directly but instead use \inpython{AlignmentAnalyzer.confusion\_matrix} for single alignments or \inpython{MultiAlignmentAnalyzer.confusion\_matrix} for aggregates.

\inpython{stringalign.visualize} provides alignment visualisation with more customisation compared to just using the simplified convenience method provided by \inpython{AlignmentAnalyzer.visualize}. These visualisations can be seamlessly integrated in, e.g., Jupyter notebooks for data exploration (example in \cref{sec:examples.usecase}). 

\inpython{stringalign.align} contains functionality for aligning strings and joining alignments. While this module is the backbone of Stringalign, it is lower-level, so we recommend using the \inpython{AlignmentAnalyzer} or the \inpython{MultiAlignmentAnalyzer}.

\section{Illustrative examples}\label{sec:examples}
This section contains three examples illustrating Stringalign's functionality. The first two demonstrate the potential for ambiguity when reporting error rates in practice, while the final example shows Stringalign's easy-to-use API and how researchers can use it to transparently evaluate an OCR transcription. For more examples, including demonstrations of token-specific metrics and how Stringalign can provide uncertainty estimations for such metrics by randomising alignments, see the Stringalign documentation\textsuperscript{\ref{footnote:docs}}.

\subsection{Examples showing ambiguities in CER and WER}\label{sec:examples.compare}
Inspired by \cite{neudecker2021survey}, we use two examples to show where CER and WER calculations differ across tools\footnote{These experiments do not consider token-specific metrics as the different tools define various similar, but non-comparable, metrics.}: A small synthetic example and a real-world example based on the analysis in \cite{neudecker2021survey}. For both examples, we apply the six tools discussed in  \cref{sec:related_work}: Calamari, Dinglehopper, the ISRI Analytic Tools, Jiwer, Meeteval and ocrevalUAtion, as well as Stringalign (using a \inpython{GraphemeClusterTokenizer} and a \inpython{SplitAtWhitespaceTokenizer} with default string normalisation for CER and WER, respectively). The results are rounded to four decimal points (the accuracy reported by the ISRI Analytic Tools and ocrevalUAtion).

\subsubsection{Small synthetic example}
We applied the evaluation tools on the text strings \texttt{'That was\textbackslash nclose! \smash{\raisebox{-0.15em}{\includegraphics[width=1em,height=1em]{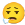}}}'} and \texttt{'That was\textbackslash nclose!'} (stored as NFC-normalised UTF-8 text files). This example is a variant of a documentation example\footnote{\url{https://stringalign.com/auto_examples/plot_emoji_ocr_evaluation.html}} and has a whitespace character different from space and a multi code point character. \Cref{tab:ToolEval} shows that different tools reported different CER and WER values. Dinglehopper and Stringalign's CER differ from the rest, as other tools count \smash{\raisebox{-0.15em}{\includegraphics[width=1em,height=1em]{bitmap.pdf}}} as three characters. Meanwhile, for WER, Dinglehopper and ISRI see no error, as they use Unicode word segmentation, which removes emojis. Jiwer only tokenises words based on \texttt{' '} and does not split at newlines, while Meeteval and Stringalign (here) splits on any space-like code point. 

\subsubsection{Real-world example}
For a multi-sample example, we applied the tools on the 378 pages from the IMPACT dataset~\cite{10.1145/2501115.2501130} analysed in \cite{neudecker2021survey}\footnote{Code to reproduce is on Stringalign's GitHub repo and the IMPACT data is available with the code for \cite{neudecker2021survey} (\url{https://github.com/cneud/hip21_ocrevaluation})}, using the same reference and predicted texts (\texttt{GT} and \texttt{GT4Hist}). \Cref{tab:ToolEval} shows the overall performance, demonstrating again that tokenisation differences causes deviations in CER and WER between tools. 
ocrevalUAtion cannot aggregate results, so for \cref{tab:ToolEval}, we extracted the CER and WER values from ocrevalUAtion's HTML reports and computed their average. This becomes a macro-average, which has the downside of weighting short and long samples equally.
Furthermore Dinglehopper's metrics are notably low, which there are two reasons for. First, it also macro-averages the results, which in this case leads to a slightly lower CER. More importantly, however, is its use of an undocumented list for resolving certain confusables. If we resolve the same list of confusables with Stringalign, then the Stringalign and Dinglehopper's CER coincide for all samples and WER coincide for most values (the overall CER results still differs due to aggregation differences). 
%Furthermore, to quantify how much the tools differed on the single-page estimates, we calculated the mean absolute deviation from the median (MADM). To compute the MADM, we, for each page, calculated the median across tools and then each tool's absolute deviation from this median. Finally, we averaged that number across all pages.
%We computed the MADM by computing each tools absolute deviation from the page median, and averaging that number across all pages.
%W%e see that Calamari, ISRI Analytic Tools and Jiwer often had values close to the median while Stringalign and ocrevalUAtion both had a larger MADM due to their tokenisation (Stringalign using grapheme clusters and ocrevalUAtion using a regular expression). Dinglehopper was an outlier, again due to its confusable handling. We see similar patterns for the WER MADM, but with larger deviations. 
Thus, consistent with the previous example, these results show that different tools do different processing under the hood (which is not always fully documented), resulting in different CER and WER for the same string pairs.

\begin{table}[t]
    \centering
    \begin{threeparttable}
        
    \caption{Results from the examples in \cref{sec:examples.compare}}
    \begin{tabular}{l@{\hspace{3ex}}rrcrr}
\toprule
 &\multicolumn{2}{c}{Simple example} & \hspace{3ex} & \multicolumn{2}{c}{IMPACT Data} \\
 \cmidrule{2-3} \cmidrule{5-6}
 %& & &  & \multicolumn{2}{c}{Overall measure}  \\
  \cmidrule{5-6}
 & CER & WER && CER & WER \\
\midrule
Calamari & 0.2105 & -- && 0.2133 & -- \\
Dinglehopper & 0.1176 & 0.0000 && 0.1821 & 0.3445 \\
ISRI & 0.2105 & 0.0000 && 0.2132 & 0.4547 \\
Jiwer & 0.2105 & 0.3333 && 0.2135 & 0.5117 \\
Meeteval & -- & 0.2500 && -- & 0.4903 \\
ocrevalUAtion & --\tnote{*} & --\tnote{*} && 0.2261\tnote{+} & 0.4675\tnote{+}  \\
Stringalign & 0.1176 & 0.2500 && 0.2120 & 0.4903\\
\bottomrule
    \end{tabular}
    \begin{tablenotes}
        \item[*] ocrevalUAtion crashed with the emoji input.
        \item[+] Manually computed macro average.
    \end{tablenotes}
    \label{tab:ToolEval}
    \end{threeparttable}
    
\end{table}

\subsection{Example use-case}\label{sec:examples.usecase}
Here, we include an abbreviated version of the tutorial example in Stringalign's documentation\footnote{\url{https://stringalign.com/worked_example.html}}. In particular, we use Stringalign to investigate the handwritten text recognition (HTR) model \texttt{Sprakbanken/TrOCR-norhand-v3}'s performance on the test split of \texttt{Teklia/NorHand-v3-line} (both pulled from Huggingface).

\Cref{fig:StringalignExample,fig:StringalignExampleVisuals} illustrate how Stringalign makes it straightforward to inspect transcription errors. The code is run in a Jupyter Notebook, and consists of five steps: First, we import modules and load data, before creating a \inpython{MultiAlignmentAnalyzer}, explicitly stating how to tokenise strings. Next, we compute the full dataset WER and list the most common edits. From these results, we see that the model struggles with the name ``Ivar'' and  replaces ``jej'' and ``eg'', which both mean the same as ``jeg'' (``I'' in Norwegian Bokmål). This requires further investigation, but could indicate that the model has overfitted to the Bokmål form of Norwegian and is struggling with other dialects. 

Finally, we see how we can use Stringalign's visualisation utilities to inspect errors more closely by displaying alignments next to the corresponding images (this requires the optional dependency \texttt{Pillow}). We see that the letter ``I'' in ``Ivar'' can resemble an ``S''. Also, the model appears to struggle with printed text and including more printed text in the training could mitigate this problem. 

Thus, Stringalign's API makes it straightforward to investigate errors made by a transcription model. This code is only parts of one example from String\-align's vast gallery that researchers can copy and learn from.

\begin{figure}
    \centering
\begin{python}    
# Import functionality and load transcriptions
import pandas as pd
from IPython.display import HTML, display
import stringalign as sa
from stringalign.evaluate import MultiAlignmentAnalyzer

data = pd.read_json("transcription.json")

# Create the alignment analyzer and compute the WER
analyzer = MultiAlignmentAnalyzer.from_strings(
    references=data["reference"],
    predictions=data["predicted"],
    metadata=[{"im_path": r.img} for r in data.itertuples()],
    tokenizer=sa.tokenize.UnicodeWordTokenizer(),
)
print(f"The WER is: {analyzer.compute_ter():.2%}")

# Print the three most common edit operations
for edit, count in analyzer.edit_counts["raw"].most_common(3):
    print(f"Edit '{edit}' occured {count} times.")

# Ivar is one of the most commonly missed words
for aa in analyzer.false_negative_index["Ivar"]:
    img = sa.visualize.create_html_image(aa.metadata["im_path"])
    alignment_html = aa.visualize(space_alignment_ops=True)
    display(HTML(img + alignment_html))
    
# Look at the three lines with the lowest WER
alignment_analyzers = sorted(
    analyzer.alignment_analyzers,
    key=lambda aa: -aa.compute_ter(),
)

for aa in alignment_analyzers[:5]:
    img = sa.visualize.create_html_image(aa.metadata["im_path"])
    alignment_html = aa.visualize(space_alignment_ops=True)
    display(HTML(img + alignment_html))
\end{python}
    \begin{monospace}
    \raggedright
    The WER is: 15.05\%\\
    Edit 'REPLACED 'jej' -> 'jeg'' occured 7 times.\\
    Edit 'REPLACED 'Ivar' -> 'Svar'' occured 7 times.\\
    Edit 'REPLACED 'eg' -> 'og'' occured 7 times.\\
    Edit 'REPLACED 'De' -> 'de'' occured 4 times.\\
    Edit 'DELETED  '1'' occured 3 times.
    \end{monospace}
    \caption{Code example with text output showing how Stringalign can help in inspecting the performance of a handwriting text recognition model. Visualisations generated by this code are shown in \cref{fig:StringalignExampleVisuals}.}
    \label{fig:StringalignExample}
\end{figure}

\begin{figure}[t]
    \centering
    \begin{subfigure}[b]{0.45\textwidth}
        \includegraphics[width=\textwidth]{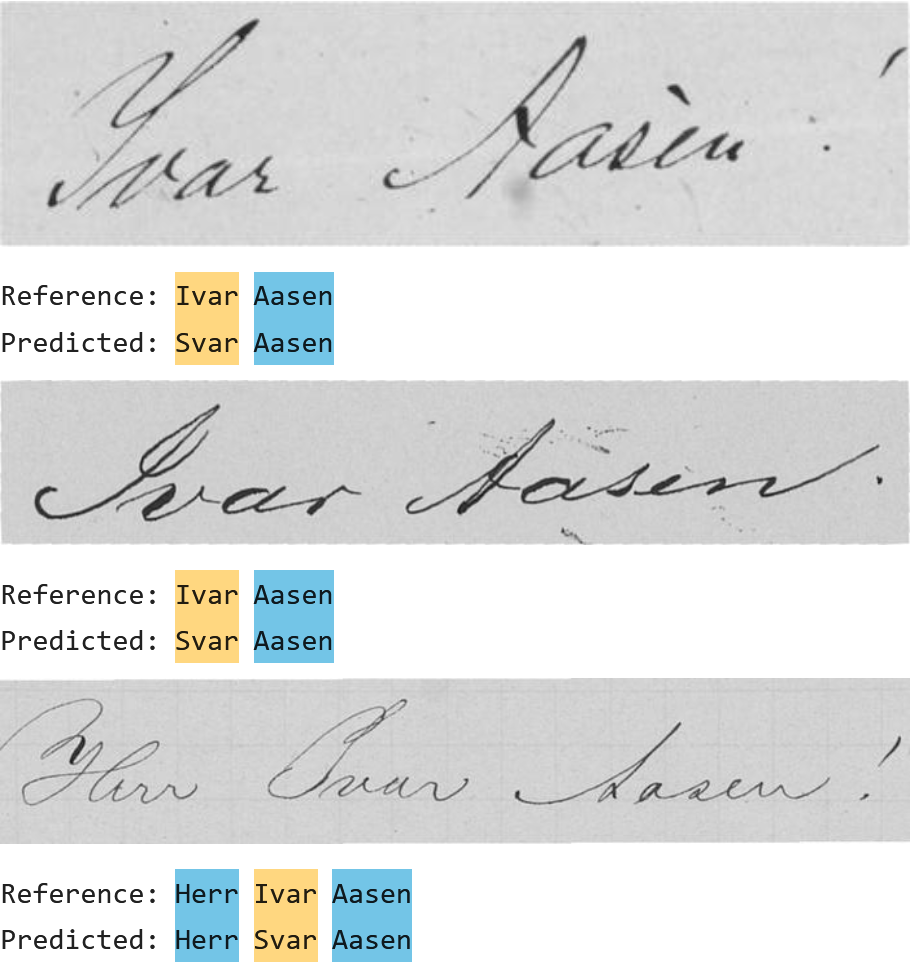}
        \caption{}
    \label{fig:StringalignExampleVisualsIvar}
    \end{subfigure}
    \begin{subfigure}[b]{0.45\textwidth}
        \includegraphics[width=\textwidth]{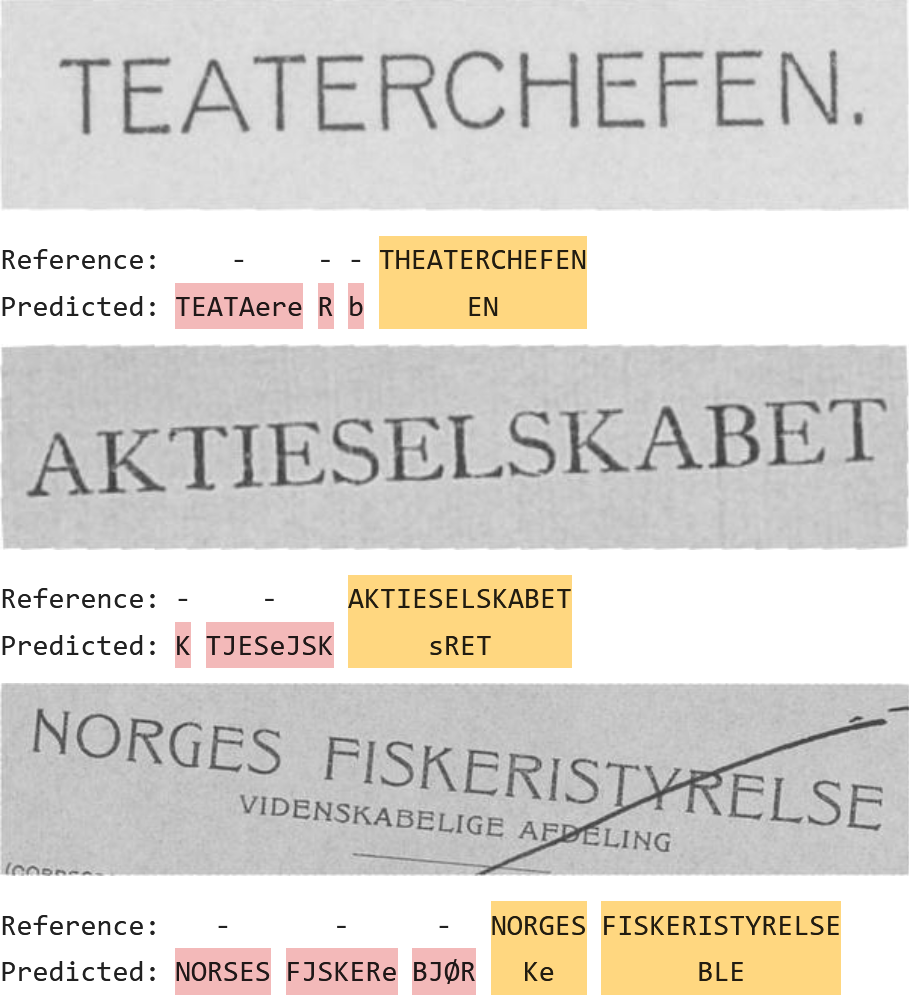}
        \caption{}
    \label{fig:StringalignExampleVisualsWER}
    \end{subfigure}
    \caption{The visual output of \cref{fig:StringalignExample} (abbreviated). (a) shows the first three lines with a false positive ``Ivar'' and (b) shows the three lines with highest WER.}
    \label{fig:StringalignExampleVisuals}
\end{figure}
\section{Conclusion and future work}\label{sec:conclusion}
This paper introduces Stringalign, an open-source Python library for comparing strings. Our work shows that where other tools can give ambiguous results, Stringalign improves on the current trancription model evaluation landscape with a transparent API, detailed documentation and tools for exploring the types of mistakes a transcription model makes, which in turn facilitates reproducible evaluation. Moreover, Stringalign has an open-source license, requires few dependencies and is modular in design, which makes it painless to adapt into a variety of existing workflows. 

This work focuses on transparent string comparison, one of the two challenges of ATR evaluation highlighted in~\cite{neudecker2021survey}. As future work, Stringalign could be further extended. For example, while performance bottlenecks are implemented in Rust, algorithmic improvements could increase speed~\cite{UKKONEN1985100}. Furthermore, the author's are only familiar with latin scripts, so this work has not focused on non-Latin scripts. Another valuable extension would, therefore, be adding support and examples for languages requiring custom tokenisation (and in particular languages using non-Latin scripts). As Stringalign is open-source, we invite the research community to provide feedback and code.

\section*{\normalsize{Disclosure of Interests}}
The authors have no competing interests to declare that are
relevant to the content of this article

\bibliographystyle{splncs04}
\bibliography{bibliography}
\end{document}